# A Self-Supervised Terrain Roughness Estimator for Off-Road Autonomous Driving

**David Stavens and Sebastian Thrun**
Stanford Artificial Intelligence Laboratory
Computer Science Department
Stanford, CA 94305-9010
{stavens,thrun}@robotics.stanford.edu

## Abstract

Accurate perception is a principal challenge of autonomous off-road driving. Perceptive technologies generally focus on obstacle avoidance. However, at high speed, terrain roughness is also important to control shock the vehicle experiences. The accuracy required to detect rough terrain is significantly greater than that necessary for obstacle avoidance.

We present a self-supervised machine learning approach for estimating terrain roughness from laser range data. Our approach compares sets of nearby surface points acquired with a laser. This comparison is challenging due to uncertainty. For example, at range, laser readings may be so sparse that significant information about the surface is missing. Also, a high degree of precision is required when projecting laser readings. This precision may be unavailable due to latency or error in pose estimation. We model these sources of error as a multivariate polynomial. The coefficients of this polynomial are obtained through a self-supervised learning process. The "labels" of terrain roughness are automatically generated from actual shock, measured when driving over the target terrain. In this way, the approach provides its own training labels. It "transfers" the ability to measure terrain roughness from the vehicle's inertial sensors to its range sensors. Thus, the vehicle can slow before hitting rough terrain.

Our experiments use data from the 2005 DARPA Grand Challenge. We find our approach is substantially more effective at identifying rough surfaces and assuring vehicle safety than previous methods – often by as much as 50%.

## 1 INTRODUCTION

In robotic autonomous off-road driving, the primary perceptual problem is terrain assessment in front of the robot. For example, in the 2005 DARPA Grand Challenge (DARPA, 2004), a robot competition organized by the U.S. Government, robots had to identify drivable surface while avoiding a myriad of obstacles – cliffs, berms, rocks, fence posts. To perform terrain assessment, it is common practice to endow vehicles with forward-pointed range sensors. Terrain is then analyzed for potential obstacles. The result is used to adjust the direction of vehicle motion (Kelly & Stentz, 1998a; Kelly & Stentz, 1998b; Langer *et al.*, 1994; Urmson *et al.*, 2004).

When driving at high speed – as in the DARPA Grand Challenge – terrain roughness must also dictate vehicle behavior because rough terrain induces shock proportional to vehicle speed. The effect of shock can be detrimental (Brooks & Iagnemma, 2005). To be safe, a vehicle must sense terrain roughness and slow accordingly. The accuracy needed for assessing terrain roughness exceeds that required for obstacle finding by a substantial margin – rough patches are often just a few centimeters in height. This makes design of a competent terrain assessment function difficult.

In this paper, we present a method that enables a vehicle to acquire a competent roughness estimator for high speed navigation. Our method uses self-supervised machine learning. This allows the vehicle to learn to detect rough terrain while in motion and without human training input. Training data is obtained by a filtered analysis of inertial data acquired at the vehicle core. This data is used to train (in a supervised fashion) a classifier that predicts terrain roughness from laser data. In this way, the learning approach "transfers" the capability to sense rough terrain from inertial sensors to environment sensors. The resulting module detects rough terrain in advance, allowing the vehicle to slow. Thus, the vehicle avoids high shock that would otherwise cause damage.

We evaluate our method using data acquired in the 2005 DARPA Grand Challenge with the vehicle shown in Fig. 1. Our experiments measure ability to predict shock and effect of such predictions on vehicle safety.

We find our method is more effective at identifying rough surfaces than previous techniques derived from obstacle avoidance algorithms. The self-supervised approach – whose training data emphasizes the distinguishing characteristics between very small discontinuities – is essential to making this possible.

Furthermore, we find our method reduces vehicle shock significantly with only a small reduction in average vehicle speed. The ability to slow before hitting rough terrain is in stark contrast to the speed controller used in the race (Thrun *et al.*, 2006b). That controller measured vehicle shock exclusively with inertial sensing. Hence, it sometimes slowed the vehicle *after* hitting rough terrain.

We present the paper in four parts. First, we define the functional form of the laser-based terrain estimator. Then we describe the exact method for generating training data from inertial measurements. Third, we train the estimator with the learning. Finally, we examine the results experimentally.

## 2  RELATED WORK

There has been extensive work on Terramechanics (Bekker, 1956; Bekker, 1969; Wong, 1989), the guidance of autonomous vehicles through rough terrain. Early work in the field relies on accurate a priori models of terrain. More recent work (Iagnemma *et al.*, 2004; Brooks & Iagnemma, 2005; Shimoda *et al.*, 2005) addresses the problem of online terrain roughness estimation. However, sensors used in that work are proprioceptive and require the robot to drive over terrain for classification. In this paper we predict roughness from laser data so the vehicle can slow in advance of hitting rough terrain.

Numerous techniques for laser perception exist in the literature, including its application to off-road driving (Shoemaker *et al.*, 1999; Urmson *et al.*, 2004). Those that address error in 3-D point cloud acquisition are especially relevant for our work. The iterative closest point (ICP) algorithm (Besl & McKay, 1992) is a well-known approach for dealing with this type of error. ICP relies on the assumption that terrain is scanned multiple times. Any overlap in multiple scans represents an error and that mismatch is used for alignment.

Although ICP can be implemented online (Rusinkiewicz & Levoy, 2001), it is not well-suited to autonomous driving. At significant range, laser data is quite sparse. Thus, although the terrain may be scanned by multiple different lasers, scans rarely cover precisely the same terrain. This effectively breaks the correspondence of ICP. Therefore, we believe that accurate recovery of a full 3-D world model from this noisy, incomplete data is impossible. Instead we use a machine learning approach to define tests that indicate when terrain is likely to be rough.

Other work uses machine learning in lieu of recovering a full 3-D model. For example, (Saxena *et al.*, 2005) and (Michels *et al.*, 2005) use learning to reason about depth in a single monocular image. Although full recovery of a world model is impossible from a single image, useful data can be extracted using appropriate learned tests. Our work has two key differences from these prior papers: the use of lasers rather than vision and the emphasis upon self-supervised rather than reinforcement learning.

In addition, other work has used machine learning in a self-supervised manner. The method is an important component of the DARPA LAGR Program. It was also used for visual road detection in the DARPA Grand Challenge (Dahlkamp *et al.*, 2006). The "trick" of using self-supervision to generate training data automatically, without the need for hand-labeling, has been adapted from this work. However, none of these approaches addresses the problem of roughness estimation and intelligent speed control of fast moving vehicles. As a result, the source of the data and the underlying mathematical models are quite different from those proposed here.

## 3  THE 3-D POINT CLOUD

Our vehicle uses a scanning laser mounted on the roof as illustrated in Fig. 2. The specific laser generates range data for 181 angular positions at 75Hz with .5 degree angular resolution. The scanning takes place in a single plane, tilted slightly downward, as indicated in the figure. In the absence of obstacles, a scanner produces a line orthogonal to the vehicle's direction of travel. Detecting obstacles, however, requires the third dimension. This is achieved through vehicle motion. As the vehicle moves, 3-D measurements are integrated over time into a 3-D point cloud.

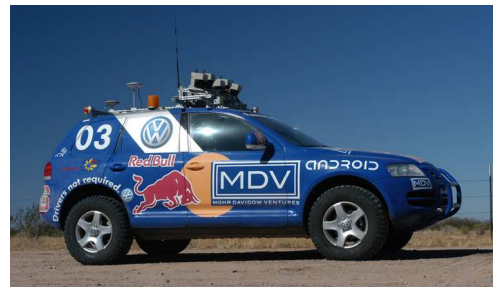

Figure 1: Stanley won the 2005 DARPA Grand Challenge by completing a 132 mile desert route in just under 7 hours. Data from this race is used to evaluate the terrain perception method described in this paper.

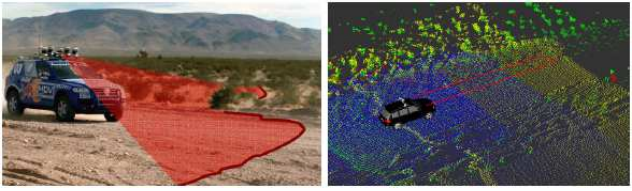

Figure 2: The left graphic shows a single laser scan. The scanning takes place in a single plane tilted downward. Reliable perception requires integration of scans over time as shown in the right graphic.

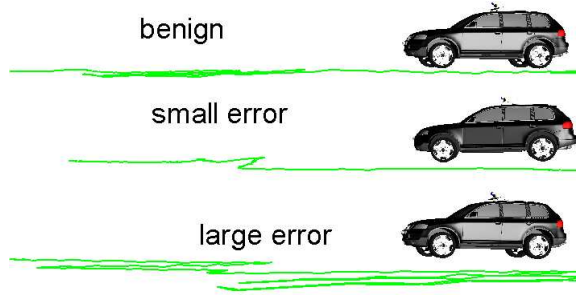

Figure 3: Pose estimation errors can generate false surfaces that are problematic for navigation. Shown here is a central laser ray traced over time, sensing a relatively flat surface.

Integration of data over time is not without problems. It requires an estimate of the coordinates at which a measurement was taken and the orientation of the sensor – in short, the *robot pose*. In Stanley, the pose is estimated from a Novatel GPS receiver, a Novatel GPS compass, an ISIS inertial measurement unit, and wheel encoder data from the vehicle. The integration is achieved through an unscented Kalman filter (Julier & Uhlmann, 1997) at an update rate of 100Hz. The pose estimation is subject to error due to measurement noise and communication latencies.

Thus, the resulting $z$-values might be misleading. We illustrate this in Fig. 3. There we show a laser ray scanning a relatively flat surface over time. The underlying pose estimation errors may be on the order of .5 degrees. When extended to the endpoint of a laser scan, the pose error can lead to $z$-errors exceeding 50 centimeters. This makes it obvious that the 3-D point cloud cannot be taken at face value. Separating the effects of measurement noise from actual surface roughness is one key challenge addressed in this work.

The error in pose estimation is not easy to model. One of the experiments in (Thrun *et al.*, 2006b) shows it tends to grow over time. That is, the more time that elapses between the acquisition of two scans, the larger the relative error. Empirical data backing up this hypothesis is shown in Fig. 4. This figure depicts the measured $z$-difference from two scans of flat terrain graphed against the time between the scans. The time-dependent characteristic suggests this be a parameter in the model.

## 4 SURFACE CLASSIFICATION

This section describes the function for evaluating 3-D point clouds. The point clouds are scored by comparing adjacent measurement points. The function for this comparison considers a number of features (including the elapsed time) which it weights using a number of parameters. The optimization of these parameters is discussed in Sect. 6.

After projection as described in Sect. 3, an individual 3-D laser point has six features relevant to our approach: its 3DOF $(x, y, z)$ position, the time $\tau$ it was observed, and first derivatives of the vehicle's estimated roll $\gamma$ and pitch $\psi$ at the time of measurement. These result in the following vector of features for each laser point $L_i$:

$$L_i = [x, y, z, \tau, \gamma, \psi] \qquad (1)$$

Our algorithm compares features pairwise, resulting in a square scoring matrix of size $N^2$ for $N$ laser points. Let us denote this matrix by $S$. In practice, $N$ points near the predicted future location of each rear wheel, separately, are used. Data from the rear wheels will be combined later (in equation (5)).

The $(r, c)$ entry in the matrix $S$ is the score generated by comparison of feature $r$ with feature $c$ where $r$ and $c$ are both in $\{0, \ldots, N-1\}$.

$$S_{r,c} = \Delta(L_r, L_c) \qquad (2)$$

The difference function $\Delta$ is symmetric and, when applied to two identical points, yields a difference of zero. Hence $S$ is symmetric with a diagonal of zeros, and its computation requires just $\frac{N^2-N}{2}$ comparisons. The element-by-element product of $S$ and the lower triangular matrix whose non-zero elements are 1 is taken. (The product zeros out the symmetric, redundant elements in the matrix.) Finally, the largest $\omega$ elements

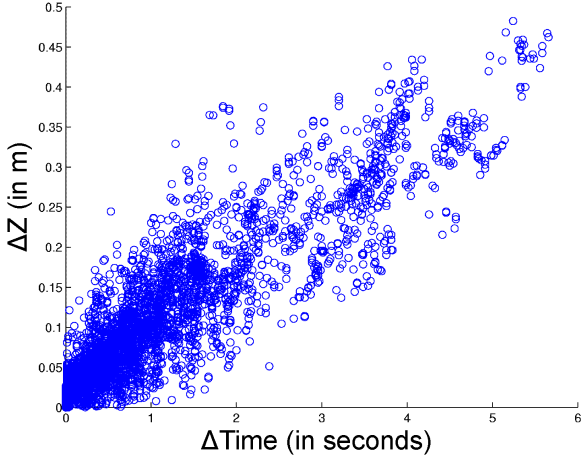

Figure 4: The measured $z$-difference from two scans of flat terrain indicates that error increases linearly with the time between scans. This empirical analysis suggests the second term in equation (4).

are extracted into the vector $W$ and accumulated in ascending order to generate the total score, $R$:

$$R = \sum_{i=0}^{\omega} W_i v^i \qquad (3)$$

Here $v$ is the increasing weight given to successive scores. Both $v$ and $\omega$ are parameters that are learned according to Sect. 6. As we will see in equation (4), points that are far apart in $(x, y)$ are penalized heavily in our method. Thus, they are unlikely to be included in the vector $W$.

Intuitively, each value in $W$ is evidence regarding the magnitude of surface roughness. The increasing weight indicates that large, compelling evidence should win out over the cumulative effect of small evidence. This is because laser points on very rough patches are likely to be a small fraction of the overall surface. $R$ is the quantitative verdict regarding the roughness of the surface.

The comparison function $\Delta$ is a polynomial that combines additively a number of criteria. In particular, we use the following function:

$$\begin{aligned} \Delta(L_r, L_c) &= \alpha_1 |L_{rz} - L_{cz}|^{\alpha_2} - \alpha_3 |L_{r\tau} - L_{c\tau}|^{\alpha_4} \\ &- \alpha_5 |euclidean(L_r, L_c)|^{\alpha_6} \\ &- \alpha_7 |L_{r\gamma}|^{\alpha_8} - \alpha_7 |L_{c\gamma}|^{\alpha_8} \\ &- \alpha_9 |L_{r\psi}|^{\alpha_{10}} - \alpha_9 |L_{c\psi}|^{\alpha_{10}} \end{aligned} \qquad (4)$$

Here $euclidean(L_i, L_j)$ denotes Euclidean distance in $(x, y)$-space between laser points $i$ and $j$. The various parameters $\alpha_{[k]}$ are generated by the machine learning.

The function $\Delta$ implicitly models the uncertainties that exist in the 3-D point cloud. Specifically, large change in $z$ raises our confidence that this pair of points is a witness for a rough surface. However, if significant time elapsed between scans, our functional form of $\Delta$ allows for the confidence to decrease. Similarly, large $(x, y)$ distance between points may decrease confidence as it dilutes the effect of the large $z$ discontinuity on surface roughness. Finally, if the first derivatives of roll and pitch are large at the time the reading was taken, the confidence may be further diminished. These various components make it possible to accommodate the various noise sources in the robot system.

Finally, we model the nonlinear transfer of shock from the left and right wheels to the IMU. (The IMU is rigidly mounted and centered just above the rear axle.) Our model assumes a transfer function of the form:

$$R_{\text{combined}} = R_{\text{left}}^{\zeta} + R_{\text{right}}^{\zeta} \qquad (5)$$

where $R_{\text{left}}$ and $R_{\text{right}}$ are calculated according to Eq. 3. Here $\zeta$ is an unknown parameter that is learned simultaneously with the roughness classifier. Clearly, this model is extremely crude. However, as we will show, it is sufficient for effective shock prediction.

Ultimately, $R_{\text{combined}}$ is the output of the classifier. In some applications, using this continuous value has advantages. However, we use a threshold $\mu$ for binary classification. (A binary assessment simplifies integration with our speed selection process.) Terrain whose $R_{\text{combined}}$ value exceeds the threshold is classified as rugged, and terrain below the threshold is assumed smooth. $\mu$ is also generated by machine learning.

## 5 DATA LABELING

The data labeling step assigns target values for $R_{\text{combined}}$ to terrain patches. This makes it possible to define a supervised learning algorithm for optimizing the parameters in our model (14 in total).

### 5.1 RAW DATA FILTERING

We record the perceived shock when driving over a terrain patch. Shock is observed with the vehicle's IMU. Specifically, the $z$-axis accelerometer is used.

However, the IMU is measuring two factors in addition to surface roughness. First, single shock events pluck resonant oscillations in the vehicle's suspension. This is problematic because we want to associate single shock events to specific, local surface data. Oscillations create an additive effect that can cause separate events to overlap making isolated interpretation difficult. Second, the gravity vector is present in the z-accelerometer data. From the perspective of the z-axis, the vector changes constantly with the pitch of the vehicle and road. This vector must be removed to generate the true shock value.

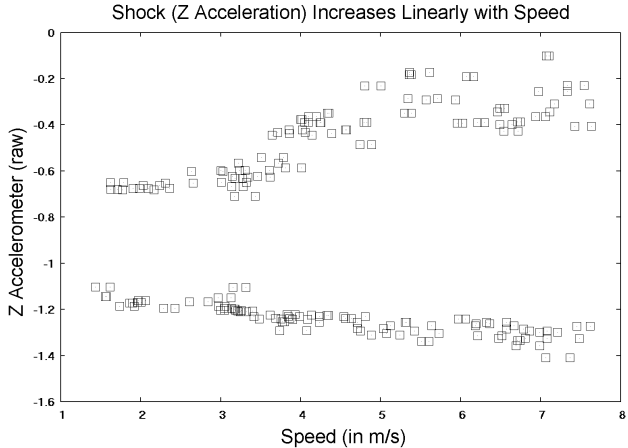

Figure 5: We find empirically that the relationship between perceived vertical acceleration and vehicle speed over a static obstacle can be approximated tightly by a linear function in the operational region of interest.

To address these issues, we filter the IMU data with a 40-tap FIR filter. The filter removes 0-10Hz from the input. Removing 0Hz addresses the gravity vector issue. Discarding the remaining frequencies addresses the vehicle's suspension.

### 5.2 CALCULATING THE SHOCK INDEX

Next, our approach calculates a series of velocity independent *ruggedness coefficients* from the measured shock values. These coefficients are used as target values for the self-supervised learning process.

This step is necessary because the raw shock filtering is strongly affected by vehicle velocity. Specifically, the faster we drive over rugged terrain, the stronger the perceived z-acceleration. Without compensating for speed, the labeling step and thus the learning breaks down. (This is especially true since data was collected with a speed controller that automatically slows down after the onset of rough terrain.)

The general dependence of perceived vertical acceleration and vehicle speed is non-linear. However, we find empirically that it can be approximated tightly by a linear function in the operational region of interest. Fig. 5 shows empirical data of Stanley traversing a static obstacle at varying speeds. The relation between shock and speed appears to be surprisingly linear. The ruggedness coefficient value, thus, is simply the quotient of the measured vertical acceleration and the vehicle speed.

## 6 LEARNING THE PARAMETERS

We now have an appropriate method to calculate target values for terrain ruggedness. Thus, we proceed to the learning method to find the parameters of the model defined in Sect. 4.

The data for training is acquired as the vehicle drives. For each sensed terrain patch, we calculate a value $R_{\text{combined}}$. When the vehicle traverses the patch, the corresponding ruggedness coefficient is calculated as above.

The objective function for learning (which we seek to maximize) is of the form:

$$T_p - \lambda F_p \qquad (6)$$

where $T_p$ and $F_p$ are the true and false positive classification rates, respectively. A true positive occurs when, given a patch of road whose ruggedness coefficient exceeds a user-selected threshold, that patch's $R_{\text{combined}}$ value exceeds $\mu$. The trade-off parameter $\lambda$ is selected arbitrarily by the user. Throughout this paper, we use $\lambda = 5$. That is, we are especially interested in learning parameters that minimize false positives in the ruggedness detection.

To maximize this objective function, the learning algorithm adapts the various parameters in our model. Specifically, the parameters are the $\alpha$ values as well as $\upsilon$, $\omega$, $\zeta$, and $\mu$. The exact learning algorithm is an instantiation of coordinate ascent. The algorithm begins with an initial guess of the parameters. This guess takes the form of a 13-dimensional vector $B$ containing the 10 $\alpha$ values as well as $\upsilon$, $\omega$, and $\zeta$. For carrying out the search in parameter space, there are two more 13-dimensional vectors $I$ and $S$. These contain an initial increment for each parameter and the current signs of the increments, respectively.

A working set of parameters $T$ is generated according to the equation:

$$T = B + SI \qquad (7)$$

The $SI$ product is element-by-element. Each element of $S$ rotates through the set {-1,1} while all other elements are held at zero. For each valuation of $S$, all values of $R_{\text{combined}}$ are considered as possible classification thresholds, $\mu$. If the parameters generate an improvement in the classification, we set $B = T$, save $\mu$, and proceed to the next element of $S$.

After a full rotation by all the elements of $S$, the elements of $I$ are halved. This continues for a fixed number of iterations. The resulting vector $B$ and the saved $\mu$ comprise the set of learned parameters.

# 7 EXPERIMENTAL RESULTS

Our algorithm was developed after the 2005 DARPA Grand Challenge, intended to replace an algorithm that adjusts speed reactively based on measured shock (Thrun *et al.*, 2006b). However, because we calculate velocity-independent ruggedness coefficients, the data recorded during the race can easily be used for training and evaluation.

Our experiments use Miles 60 to 70 of the 2005 Grand Challenge route as a training set and Miles 70 to 80 as the test set. The 10 mile-long training set contains 887,878 laser points and 11,470 shock "events" whose intensity ranges from completely negligible (.02 Gs) to very severe (nearly .5 Gs). The 10 mile test set contains 1,176,375 laser points and 13,325 shock "events". Intensity again ranges from negligible (.02 Gs) to severe (over .6 Gs). All laser points fall within 30cm of terrain the vehicle's rear tires traversed during the race. (And it is straightforward to predict the vehicle's future position.)

There is no correlation between the number of laser points in the event and its speed-normalized shock. Because the number of laser points is entirely dependent on event length (the laser acquisition rate is at constant Hz), there is also no correlation between event length and speed-normalized shock. Data from all lasers is used. Readings from lasers aimed farther are more difficult to interpret because orientation errors are magnified at range. Such lasers also produce fewer local readings because the angular resolution of the emitter is constant.

As motivated in Sect. 5.2 and Sect. 6, we select (speed-normalized) shocks of .02 Gs per mph or more as those we wish to detect. There are 42 such events in the training set (.36%) and 58 events in the test set (.44%). Because these events are extremely rare and significant uncertainty exists, they are difficult to detect reliably.

The value of .02 Gs / mph was selected because these events are large enough to warrant a change in vehicle behavior. However, they are not so rare that reliable learning is impossible.

## 7.1 SENSITIVITY VERSUS SPECIFICITY

We analyze the sensitivity versus the specificity of our method using Receiver Operating Characteristic (ROC) curves. For comparison, we also include the ROC curves for the actual terrain classification used during the race (Thrun *et al.*, 2006a). These methods are not fully comparable. The method in (Thrun *et al.*, 2006a) was used for steering decisions, not for velocity control. Further, it was trained through human driving, not by a self-supervised learning approach. Nevertheless, both methods attach numbers to points in a 3-D point cloud which roughly correspond to the ruggedness of the terrain.

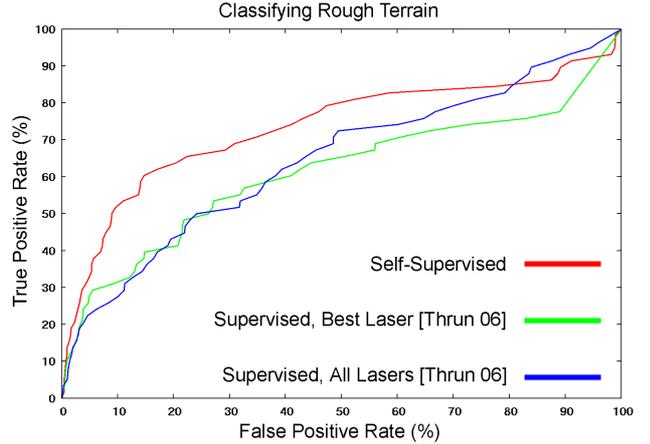

Figure 6: The true positive / false positive trade-off of our approach and two interpretations of prior work. The self-supervised method is significantly better for nearly any fixed acceptable false-positive rate.

We use the method in (Thrun *et al.*, 2006a) to generate a vector $W$. Separately trained parameters $\upsilon$, $\omega$, $\zeta$, and $\mu$ are then used to extract a prediction. Further, we acquire data for the (Thrun *et al.*, 2006a) approach in two different ways. The first considers data from all lasers equally. The second uses only data from the closest, most accurate laser.

Fig. 6 shows the resulting ROC curves. On the vertical axis, this figure plots the true-positive rate. On the horizontal axis, it plots the false-positive rate. For all of the useful false-positive ranges and more (0-80%), our approach offers significant improvement over the method in (Thrun *et al.*, 2006a). Performance in the most important region (0-20% false positive rate) is particularly good for our new approach.

At the extreme right of the plot, where the true-positive to false-positive-ratio approaches one, all the methods deteriorate to poor performance. However, within this region of poor performance, our method appears to lag behind the one in (Thrun *et al.*, 2006a). This breakdown occurs for high true positive rates because the remaining few undetected positives are areas of the surface with very high noise.

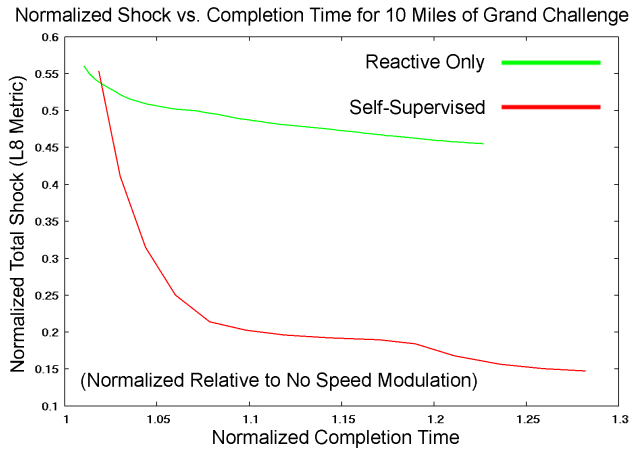

Figure 7: The trade off of completion time and shock experienced for our new self-supervised approach and the reactive method we used in the Grand Challenge.

As an aside, we notice within the (Thrun *et al.*, 2006a) approach that the most accurate laser is not strictly better than all lasers together. This is due to the trade-off between resolution and noise. The closest laser is the least noisy. However, certain small obstacles are harder to detect with the lower perceptual resolution of a single laser.

## 7.2 EFFECT ON THE VEHICLE

As noted above, during the race we used a different approach for velocity control (Thrun *et al.*, 2006b). The vehicle slowed after an onset of rough terrain was detected by the accelerometers. The parameters of the deceleration and the speed recovery rate were set by machine learning to match human driving. Because roughness is spatially correlated, we found this method to be quite effective. However, it did not allow the vehicle to detect rough terrain in advance. Therefore, the vehicle experienced more shock than necessary.

To understand how much improvement our new method offers, we compare it to the reactive approach used in the race. The only software change is the mechanism that triggers speed reduction. By anticipating rough terrain, our new method should allow faster finishing times with lower shock than previously possible.

This is indeed the case. The results are shown in Fig. 7. The graph plots, on the vertical axis, the shock experienced. The horizontal axis indicates the completion time of the 10-mile section used for testing. Both axes are normalized by the values obtained with no speed modification. The curve labeled "reactive only" corresponds to the actual race software. The curve labeled "self-supervised" indicates our new method.

The results clearly demonstrate that reducing speed proactively with our new terrain assessment technique constitutes a substantial improvement. The shock experienced is dramatically lower for essentially all possible completion times. The reduction is as much as 50%. Thus, while the old method was good enough to win the race, our new approach would have permitted Stanley to drive significantly faster, still avoiding excessive shock due to rough terrain.

## 8 DISCUSSION

We have presented a novel, self-supervised learning approach for estimating the roughness of outdoor terrain. Our main application is the detection of small discontinuities likely to create significant shock for a high speed robotic vehicle. By slowing, the vehicle can reduce the shock it experiences. Estimating roughness demands the detection of very small surface discontinuities – often a few centimeters. Thus the problem is significantly more challenging than finding obstacles.

Our approach monitors the actual vertical accelerations caused by the unevenness of the ground. From that, it extracts a ruggedness coefficient. A self-supervised learning procedure is then used to predict this coefficient from laser data, using a forward-pointed laser. In this way, the vehicle can safely slow down *before* hitting rough surfaces. The approach in this paper was formulated (and implemented) as an offline learning method. But it can equally be used as an online learning method where the vehicle learns as it drives.

Experimental results indicate that our approach offers significant improvement in shock detection and – when used for speed control – reduction. Compared to prior methods, our algorithm has a significantly higher true-positive rate for (nearly) any fixed value of acceptable false-positive rate. Results also indicate our new proactive method allows for additional shock reduction without increase in completion time compared to our previous reactive work. Put differently, Stanley could have completed the race even faster without any additional shock.

There exist many open questions that warrant further research. One is to train a camera system so rough terrain can be extracted at farther ranges. Also, the mathematical model for surface analysis is quite crude. Further work could improve upon it. Additional parameters, such as the orientation of the vehicle and the orientation of the surface roughness, might improve performance. Finally, we believe using our method online, during driving, could add insight to its strengths and weaknesses.

**Acknowledgments**

David Stavens' doctoral study is supported by the David Cheriton Stanford Graduate Fellowship (SGF). Gabe Hoffmann of the Stanford Racing Team designed the 40-tap FIR filter used for extracting vehicle shock which is gratefully acknowledged.